\documentclass{article}

    \usepackage[preprint]{neurips_2022}

\usepackage[utf8]{inputenc} 
\usepackage[T1]{fontenc}    
\usepackage{hyperref}       
\usepackage{url}            
\usepackage{booktabs}       
\usepackage{amsfonts}       
\usepackage{nicefrac}       
\usepackage{microtype}      
\usepackage{xcolor}         
\usepackage{natbib}
\usepackage{graphicx}
\usepackage{threeparttable}
\usepackage{multirow}
\usepackage{bm}
\usepackage{amsmath}
\usepackage{ulem}

\DeclareUnicodeCharacter{2061}{}
\title{Mol-PECO: a deep learning model to predict human olfactory perception from molecular structures}

\author{Author One\qquad Author Two\\Affiliation}

\author{
  Mengji Zhang\thanks{This work is done as a visiting student in Tokyo University.} \\
  Schoool of Biomedical Engineering, Shanghai Jiao Tong University, Shanghai, 200030, China \\
  Institute of Industrial Science, The University of Tokyo, Meguro-ku, Tokyo, 153-8505, Japan\\
  \texttt{mengji.zhang0809@gmail.com} \\
  \And
  Yusuke Hiki\\
  Department of Biosciences and Informatics\\
  Keio University\\
  Kouhoku-ku, Yokohama, 223-8522, Japan\\
  \texttt{hiki@fun.bio.keio.ac.jp}\\  
  \And
  Akira Funahashi\\
  Department of Biosciences and Informatics\\
  Keio University\\
  Kouhoku-ku, Yokohama, 223-8522, Japan\\
  \texttt{funa@bio.keio.ac.jp}\\
  \And
  Tetsuya J. Kobayashi\thanks{Correspondence to Tetsuya J. Kobayashi or Mengji Zhang.} \\
  Institute of Industrial Science \\
  The University of Tokyo \\
  Meguro-ku, Tokyo, 153-8505, Japan\\
  \texttt{tetsuya@mail.crmind.net} \\
}

\begin{document}

\maketitle



\begin{abstract}
While visual and auditory information conveyed by wavelength of light and frequency of sound have been decoded, predicting olfactory information encoded by the combination of odorants remains challenging due to the unknown and potentially discontinuous perceptual space of smells and odorants. Herein, we develop a deep learning model called Mol-PECO (\textbf{Mol}ecular Representation by \textbf{P}ositional \textbf{E}ncoding of \textbf{Co}ulomb Matrix) to predict olfactory perception from molecular structures. Mol-PECO updates the learned atom embedding by directional graph convolutional networks (GCN), which model the Laplacian eigenfunctions as positional encoding, and Coulomb matrix, which encodes atomic coordinates and charges. With a comprehensive dataset of $8,503$ molecules, Mol-PECO directly achieves an area-under-the-receiver-operating-characteristic (AUROC) of $0.813$ in $118$ odor descriptors, superior to the machine learning of molecular fingerprints (AUROC of $0.761$) and GCN of adjacency matrix (AUROC of $0.678$). The learned embeddings by Mol-PECO also capture a meaningful odor space with global clustering of descriptors and local retrieval of similar odorants. 
Our work may promote the understanding and decoding of the olfactory sense and mechanisms.
\end{abstract}

\section{Introduction}
Olfaction is one of the essential senses, where the sense of smell is triggered by the binding of odorant molecules to olfactory receptors and is shaped by the subsequent neural processing of the received information in the brain\citep{sobel1998sniffing, lapid2011neural}. 
Unlike vision and hearing, however, the prediction of olfactory perception for odorant molecules remains challenging. 
On the one hand, some molecules with different functional groups share identical smells\citep{sell2006unpredictability}. 
On the other hand, other molecules with similar structures can produce totally different perceptions\citep{boesveldt2010carbon}.
The structures and perceptions of the odorant molecules are nonlinearly and discontinuously related\citep{keller2017predicting}.
Unvieling the quantitative structure-odor relationship (QSOR) is indispensable for understanding the coding principle of olfactory information\citep{keller2017predicting} and also essential for predicting and designing smells and flavors for various applications such as food technologies\citep{polster2017structure}. 

Machine learning (ML) is a promising approach to untangle such a complicated relationship.
However, the prediction of olfactory perception from molecular structures is strongly dependent on molecular representation\citep{pattanaik2020molecular}. 
Molecular fingerprints, which encode chemical substructures into fixed-length vectors, are the major and classical molecular representation, yet demonstrate limited performances in QSOR due to the inefficient feature extraction by hand-crafted rules\citep{rogers2010extended, moriwaki2018mordred}. 
To learn a good representation of molecules from data, graph convolutional networks (GCNs) have been widely applied in molecular modeling\citep{kipf2016semi}, e.g., quantum chemistry\citep{hofstetter2022graph, yang2019analyzing, feinberg2018potentialnet}, biophysics\citep{wang2019molecule, withnall2020building}, and biological side effects\citep{yang2019analyzing, withnall2020building}. 
The conventional GCN models each molecule by the adjacency matrix, which encodes the chemical bonds as a graph, and performs the information aggregation among the neighbors prescribed by the adjacency matrix\citep{yang2019analyzing, sanchez2019machine}.
While GCN has been reported to outperform conventional ML with molecular fingerprints in standard tasks\citep{sanchez2019machine}, it still has technical drawbacks, which potentially limit the applicability of GCN and the adjacency matrix (\citep{kreuzer2021rethinking, oono2019graph, topping2021understanding}) to learning QSOR.
First, the adjacency matrix cannot encode the atomic and global 3D information of molecules\citep{pattanaik2020molecular}, even though such atomic and 3D information is the major determinant of binding affinities between odorant molecules and olfactory receptors\citep{floriano2000molecular}.
Second, graphs do not have a canonical coordinate representation, which contrasts with sequences and images for which one-, two-, and three-dimensional lattice coordinates are canonical.
As a result, the GCN employs permutation invariant operations, e.g., message passing or neighboring aggregation, which then limit its expressive power to discriminate molecules with different structures(\citep{kreuzer2021rethinking}) and also induces oversmoothing\citep{oono2019graph} and oversquashing\citep{topping2021understanding}.
All of these factors may hamper the network to efficiently learn the QSOR of odorant molecules with a variety of structures.

\begin{figure}[htbp]
\centering
\includegraphics[width=0.95\linewidth]{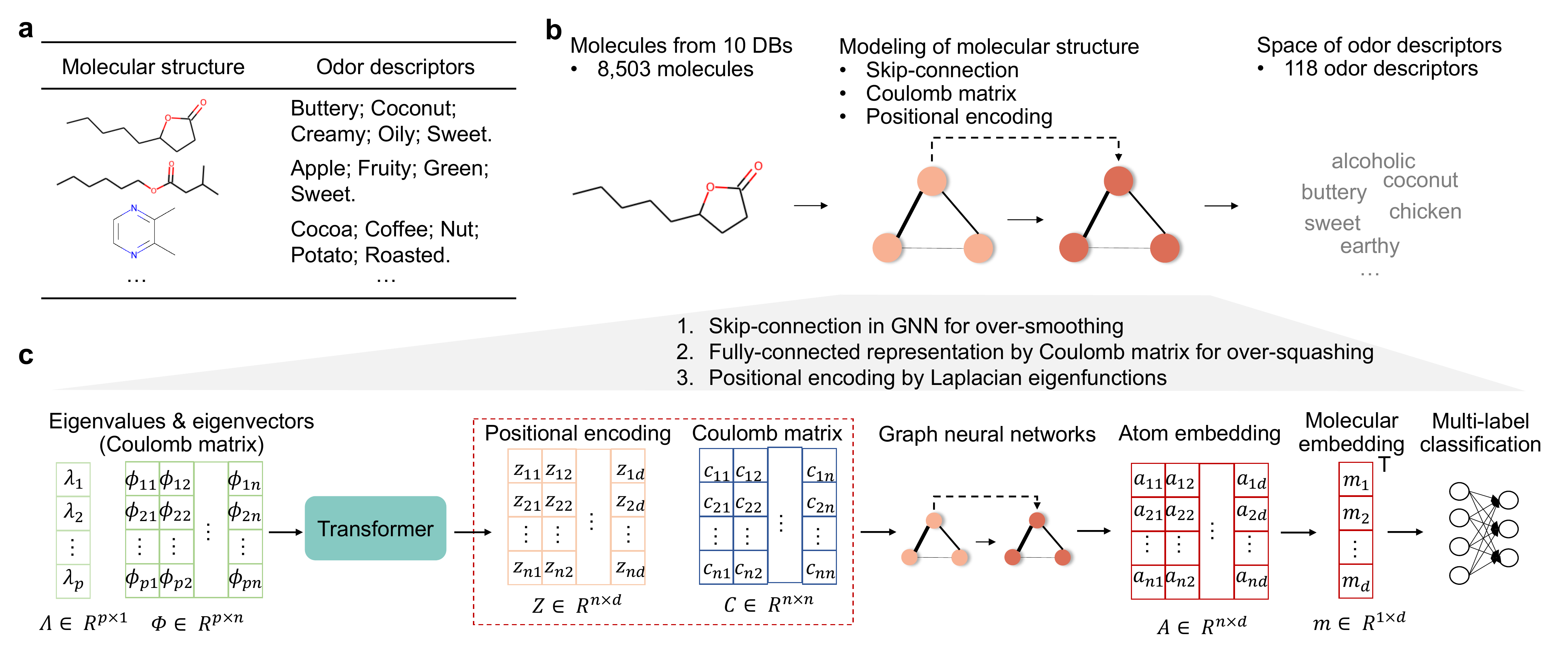}
\caption{Overview of Mol-PECO. (a) Typical molecular structures and the corresponding odor descriptors are shown as examples. (b) The main workflow of modeling quantitative structure-odor relationship (QSOR). (c) The detailed model architecture of Mol-PECO and its three features: 1) skip-connection in graph neural networks to alleviate over-smoothing, 2) fully-connected molecular representation by Coulomb matrix to suppress  over-squashing, and 3) positional encoding by Laplacian eigenfunctions.}
\label{fig:fig1}
\end{figure}

In this work, we develop a deep learning model (Mol-PECO) for QSOR, which aims at the multi-label classification of olfactory perception from molecular structure (Figure \ref{fig:fig1}).
To address the problems in conventional ML and GCN approaches, Mol-PECO combines the Coulomb matrix (CM) and Spectral Attention Network (SAN).
CM is a simple global representation of a molecule by Coulombic forces between atoms in the molecule calculated with the nuclear charges and the corresponding 3D coordinates\citep{rupp2012fast}.
CM therefore could encode more detailed structural information than the adjacency matrix which only represents the chemical bonds between atoms.  
SAN is an architecture of graph attention network (GAT)\citep{kreuzer2021rethinking}, which uses the full Laplacian spectrum of a molecular graph for a learned positional encoding (LPE).
Eigenfunctions of the graph Laplacian hierarchically describe the global and local structures in a graph and thereby provides a way to canonically characterize graphs and to define positional information of nodes (atoms)\citep{dwivedi2020generalization}.
SAN can also be categorized as an attempt to extend the expressive power in graph-based architectures by endowing a canonical coordinate or positional information with graphs by using their spectral information\citep{beaini2021directional, kreuzer2021rethinking,dwivedi2020generalization}.
CM can be naturally combined with SAT by regarding CM as a weighted adjacency matrix. 

Based on the learned representation, Mol-PECO directly predicts $118$ odor descriptors of perception for each odorant molecule. 
Mol-PECO achieves area under the receiver operating characteristic curve (AUROC) of $0.813$ and area under the precision-recall curve (AUPRC) of $0.181$, whereas the conventional MLs of molecular fingerprints fail to balance AUROC and AUPRC; the ML method (cfps-KNN) with the highest AUROC of $0.761$ has low AUPRC of $0.057$ and one (mordreds-RF) with the highest AUPRC of $0.144$ shows low AUROC of $0.723$.
Thus, Mol-PECO may boost the prediction of QSOR for applications and also contribute to the understanding of the principle underlying olfactory information processing.

\section{Dataset:A comprehensive human olfactory perception dataset}
In this work, we use the data in which each molecular structure is paired with multiple odor descriptors (Figure \ref{fig:fig1} a). 
The dataset in our work is compiled from ten expert-labeled sources: Arctander's dataset ($n = 3,102$, \citep{arctander1960perfume}), AromaDb ($n = 1,194$, \citep{kumar2018aromadb}), FlavorDb ($n = 525$, \citep{garg2018flavordb}), FlavorNet ($n = 718$, \citep{acree2004flavornet}), Goodscents ($n = 6,158$, \citep{flavorfood}), Fragrance Ingredient Glossary ($n = 1,135$, \citep{international2003gc}), Leffingwell's dataset ($n = 3,523$, \citep{leffingwell2001olfaction}), Sharma's dataset ($n = 4,006$, \citep{sharma2021smiles}), OlfactionBase ($n = 5,105$, \citep{sharma2022olfactionbase}), and Sigma's Fragrance and Flavor Catalog ($n = 871$, \citep{sigma2011aldrich}). 
These datasets are retrieved from the archive of \url{https://github.com/pyrfume/pyrfume-data}.
The data cleaning procedure includes 1) merging the overlapped molecules, 2) filtering the conflict descriptors, and 3) filtering the rare descriptors assigned to less than $30$ molecules. After data cleaning, we obtain a comprehensive dataset of $8,503$ molecules and $118$ odor descriptors.

\begin{figure}[htbp]
\centering
\includegraphics[width=0.95\linewidth]{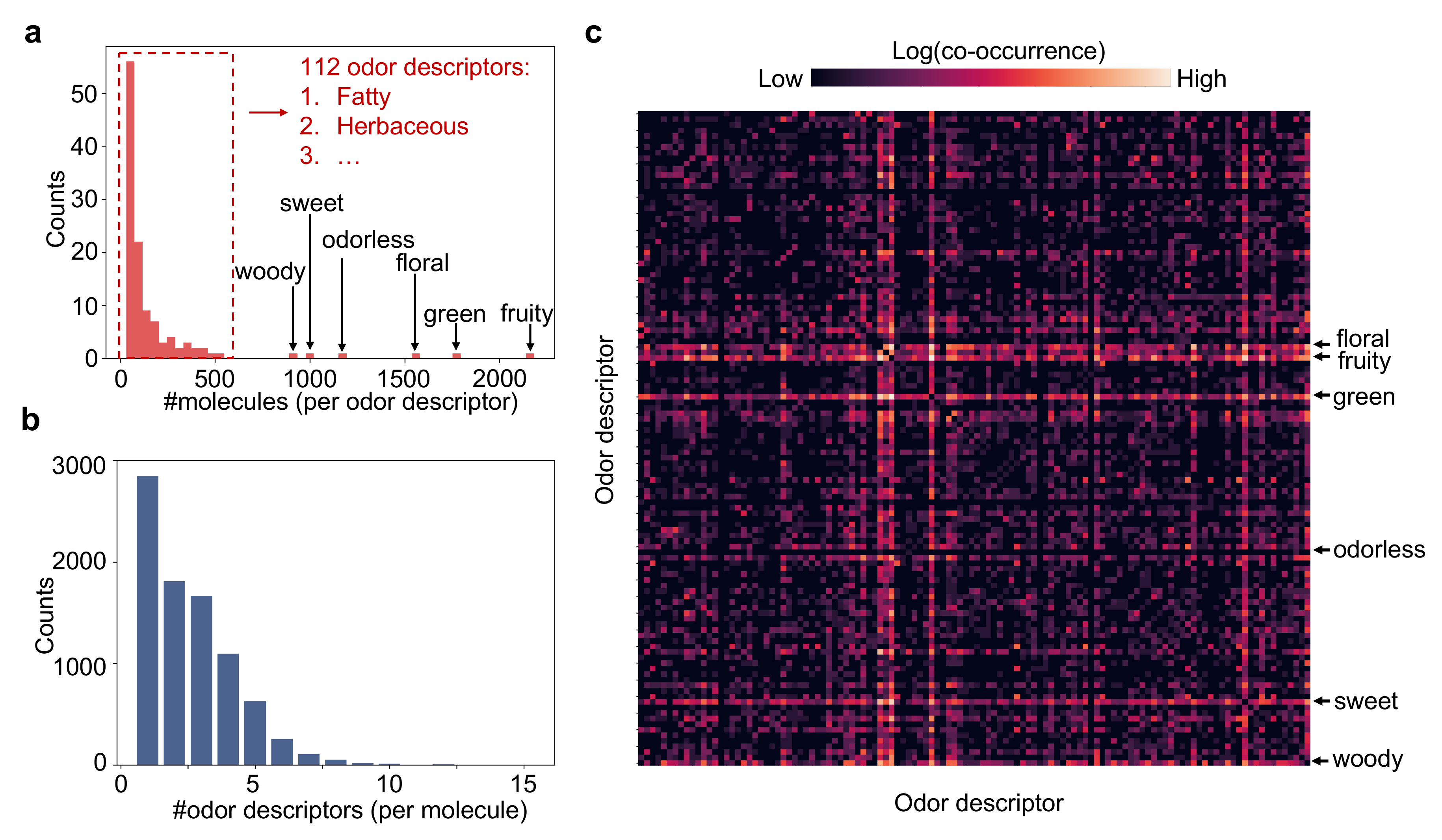}
\caption{The comprehensive human olfactory dataset. (a) Distribution of molecules across odor descriptors. (b) Distribution of descriptors across molecules. (c) Co-occurrence matrix of 118 odor descriptors.  The heatmap is demonstrated with logarithm transformation, and the descriptors are ordered alphabetically.}
\label{fig:fig2}
\end{figure}

This comprehensive human olfactory perception dataset is multi-labeled, with every molecule labeled with one or several odor descriptors. 
For the number of molecules associated with each odor descriptor, the distribution is imbalanced: each of $112$ odor descriptors possesses $\leq 800$ molecules whereas the other $6$ descriptors possess $> 800$ molecules (Figure \ref{fig:fig2} a). 
For the number of descriptors associated with each molecule, the distribution is also skewed, with $8,054$ molecules possessing $\leq 5$ odor descriptors and $449$ molecules possessing $> 5$ odor descriptors (Figure \ref{fig:fig2} b). For co-occurrence, descriptors of 'fruity', 'green', 'sweet', 'floral', and 'woody' co-occur with almost all the descriptors, while odorless molecules co-occur with no other molecules (Figure \ref{fig:fig2} c). 
The data split is built by second-order iterative stratification \citep{szymanski2017network}, which aims at splitting multi-label dataset and preserves the label ratios in each split with an iterative sampling design. 
The whole dataset is splitted into train / validation / test datasets of $6,802$ / $864$ / $837$ pairs, respectively.

\section{Results}
In this section, we first introduce Coulomb-GCN, which updates GCN by replacing the adjacency matrix with CM. After verifying the effectiveness of Coulomb-GCN, we have Mol-PECO by further replacing the random embedding of atoms in Coulomb-GCN with positional encoding, in which the spectral information of the CM is employed to have a structure-aware embedding.

\subsection{Fully-connected graph by Coulomb matrix is superior to sparse graph by adjacency matrix}
\begin{figure}[htbp]
\centering
\includegraphics[width=0.95\linewidth]{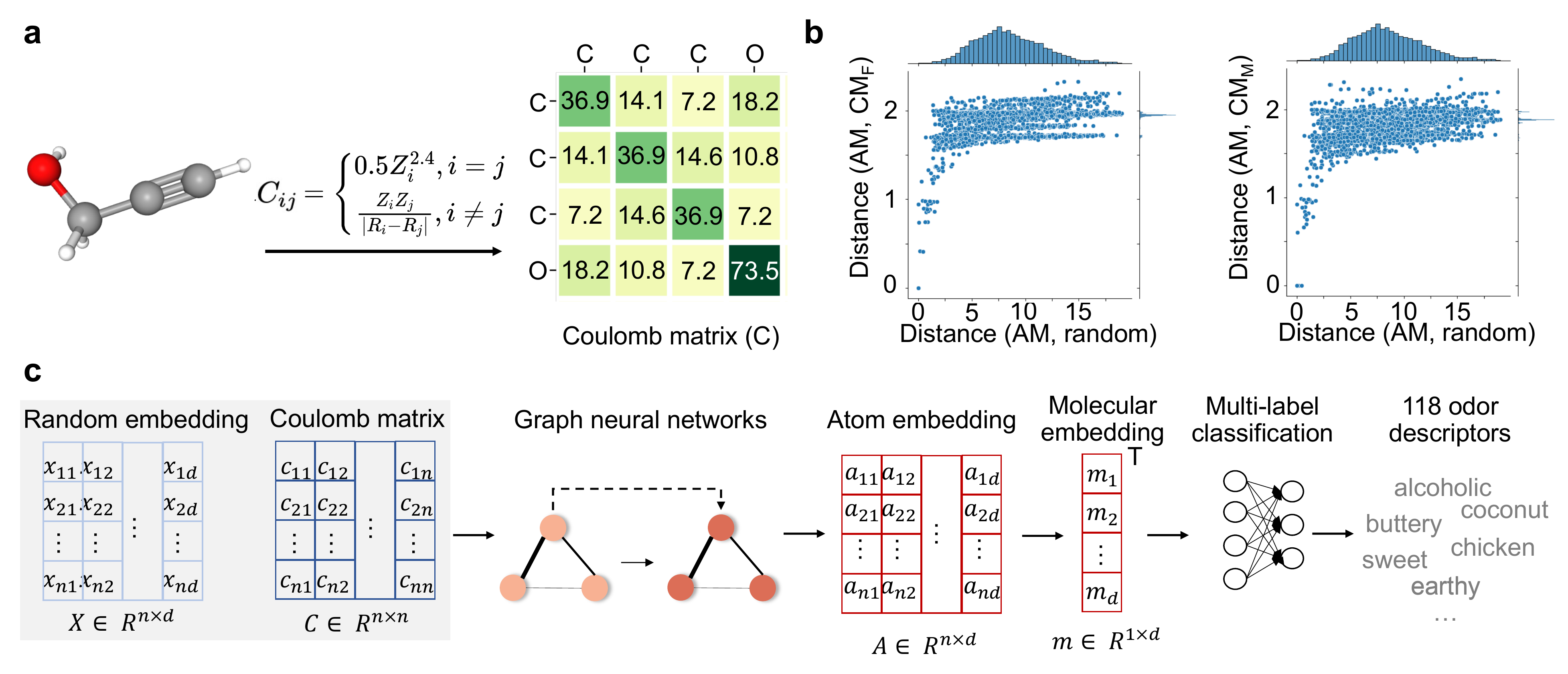}
\caption{The motivation and workflow of modeling the Coulomb matrix. (a) An example of Coulomb matrix as a fully-connected graph for Propargyl alcohol, a clear colorless liquid with a geranium-like odor. (b) Similarity between Coulomb matrix and adjacency matrix, indicated by the distances. The distance is calculated with the Frobenius norm. (c) The workflow of modeling Coulomb matrix by graph neural networks.}
\label{fig:fig3}
\end{figure}

We calculate CM, which models atomic energies with the internuclear Coulomb repulsion operator \citep{rupp2012fast, schrier2020can}, and use it as our molecular representation. 
In CM, the diagonal entries refer to a polynomial fit of atomic energies and off-diagonal entries represent the Coulomb repulsion force between atomic nuclei. 
Although the adjacency matrix has been widely used in molecular modeling\citep{mahmood2021masked, wang2022molecular}, CM as an emerging molecular representation can have at least two advantages: 1) CM handles the over-squashing plight by allowing direct paths between distant nodes in the fully-connected graph representation (Figure \ref{fig:fig3}a); 2) distance by Frobenius norm between CM and adjacency matrix is $5$--$10$ times smaller than that between random initialized matrix and adjacency matrix, indicating that CM is fully-connected while preserving a similarity to adjacency matrix (Figure \ref{fig:fig3}b).


We build a nonlinear map (named Coulomb-GCN) between molecular structures and human olfactory perception (Figure \ref{fig:fig3}c) by replacing the adjacency matrix in message passing of GCN with CM.
Specifically, starting from random atom embedding, the learned atom embedding is obtained by message passing on fully-connected molecular graph with neighbor weights specified by the entries of CM.
The molecular embedding is extracted by sum pooling and fed to a multi-label classification module to predict $118$ odor descriptors. 
Considering the gap between maximal and minimal entries in CM, normalization of entries may affect the performance. We test Minmax and Frobenius normalizations in a matrix-wise manner.  

We evaluate and compare the prediction accuracy of GCN with adjacency matrix and those of Coulomb-GCN with the different normalizations of CM (Table 1). 
Compared with GCN with adjacency matrix (AUROC of $0.678$), gains in AUROC are observed in Coulomb-GCN with Frobenius normalization (AUROC of $0.759$) and minmax normalization (AUROC of $0.713$).
Coulomb-GCN with Frobenius normalization also achieves higher performances in five out of six evaluation metrics (Table 1): AUROC (improved from $0.678$ to $0.759$),  AUPRC (improved from $0.111$ to $0.143$), specificity (improved from $0.625$ to $0.744$), precision (improved from $0.079$ to $0.089$), and accuracy (improved from $0.726$ to $0.780$).

\begin{table}[htbp]
\label{tab1}
\centering  
\begin{threeparttable}  
  \caption{Prediction performances of Coulomb matrix with Minmax and Frobenius normalizations and adjacency matrix in GCN. The highest score of each metric is shown in bold.}  
  {\scriptsize \normalsize
  \setlength\tabcolsep{4pt}
  \begin{tabular}{cccccccc}
  \toprule  
{Representation} & {Normalization\tnote{1}} & {AUROC} & {AUPRC} & {Precision} & {Recall} & {Specificity} & {Accuracy}\\  
  \midrule
Adjacency matrix & - & 0.678 & 0.111 & 0.079 & \textbf{0.827} & 0.625 & 0.726\\
Coulomb matrix & Minmax & 0.713 & 0.138 & 0.082 & 0.811 & 0.687 & 0.749\\
Coulomb matrix & Frobenius & \textbf{0.759} & \textbf{0.143} & \textbf{0.089} & 0.816 & \textbf{0.744} & \textbf{0.780}\\
  \bottomrule   
\end{tabular}   
\begin{tablenotes}  
\item[1] Normalization refers to the normalization methods for Coulomb matrix. 
\end{tablenotes}  
}  
\end{threeparttable}  
\end{table}

\subsection{Directional graph modeling by Laplacian eigenfunctions improves prediction accuracy}
The graph Laplacian and its spectral information enable us to characterize the global and substructures of graphs\citep{mohar1991laplacian, chung1997spectral, spielman2012spectral}. Specifically, the graph Laplacian is defined as $L = D - A$, where $D$ and $A$ refer to the degree and adjacency matrices. 
$L$ is positive semi-definite with one trivial and the other nontrivial eigenvalues. 
In this work, the Laplacian defined by CM acts as an extension of the normal Laplacian ($L = D- W$), where $W$ refers to the weighted matrix (CM in this work) and possesses the same properties as the graph Laplacian (e.g., symmetric and positive semi-definite). 
In particular, the eigenvectors of $L$ provide an optimal solution to the Laplacian quadratic form ($f^TLf = 1/2\sum_{i, j}X(i, j){(f_i-f_j)}^2$)\citep{mohar1991laplacian, spielman2012spectral, chung1997spectral}, encoding the geometric information of graphs. 

\begin{figure}[htbp]
\centering
\includegraphics[width=0.95\linewidth]{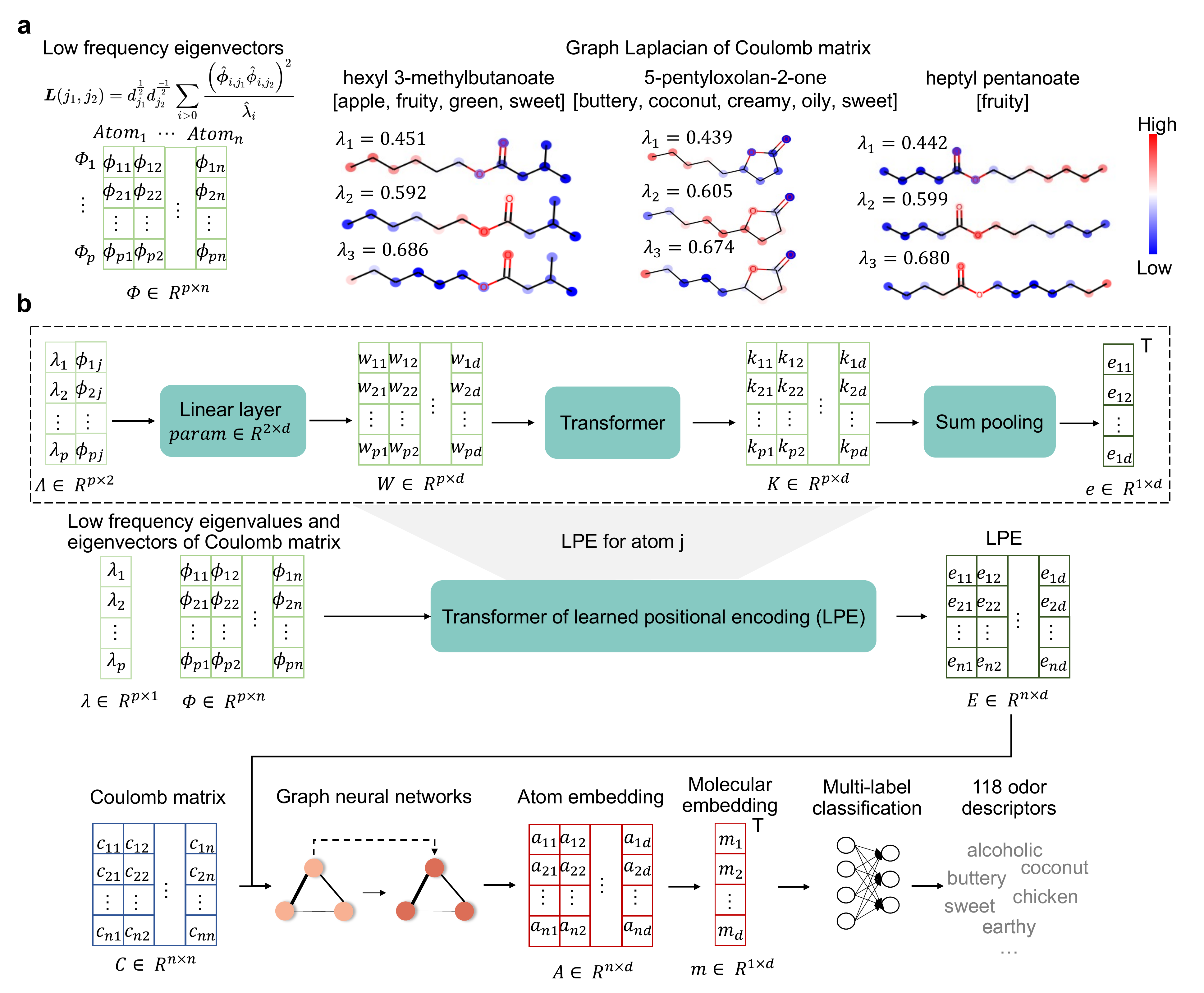}
\caption{The motivation and architecture of Mol-PECO. (a) Structural information carried by the Laplace spectrum of the Coulomb matrix. Low-frequency eigenvectors, calculated with graph Laplacian, as the input matrix for positional encoding and $3$ examples, including cyclic and acyclic molecules, of eigenvalue $\lambda_{i}$ and eigenvector $\phi_{i}$ for molecular graphs ($i \in \{1,2,3\}$). Color indicates the value of each component (node) of the eigenvectors. (b) The architecture of Mol-PECO. 
Mol-PECO learns the positional encoding (LPE) with Transformer of graph Laplacian, and updates the atom embedding with GCN of Coulomb matrix and LPE. Specifically, GCN is implemented with skip-connection to relieve over-smoothing. Coulomb matrix, the fully connected graph representation, suppresses over-squashing with direct connections between nodes. With the updated atom embedding, Mol-PECO extracts the molecular embedding with sum pooling, and predicts 118 odor descriptors with neural networks of molecular embedding. In this work, $p$ and $d$ is set as $20$ and $32$, respectively.}
\label{fig:fig4}
\end{figure}

Given these properties of the Laplacian graph, we use the Laplacian eigenfunctions of the CM to encode the positional information of molecular graphs. Typical results of a cyclic odorant (5-pentyloxolan-2-one, flowing from left to right in $\lambda_1$) and acyclic odorants (hexyl 3-methylbutanoate, flowing from left to right in $\lambda_1$, and heptyl pentanoate, flowing from right to left in $\lambda_1$) demonstrate the information carried by low frequency eigenfunctions (Figure \ref{fig:fig4} a). 
Combining it with the Coulomb-GCN, we construct the deep learning framework, named Mol-PECO (Figure \ref{fig:fig4}b), with the fully-connected molecular representation by CM and the positional encoding by Laplacian. 
We choose LPE by Transformer\citep{kreuzer2021rethinking} to build the atom embedding. Specifically, LPE concatenates the $p$ lowest eigenvalues and the corresponding eigenvectors as the input matrix $\Lambda \in R^{p \times 2}$, and learns the encoding with Transformer for every atom\citep{kreuzer2021rethinking}. We obtain AUROC of 0.796 and AUPRC of 0.153 with LPE of raw CM. We further perform the experiments for LPE of asymmetric normalized CM and obtain additional gain of performances by 0.017 and 0.028 for AUROC and AUPRC, respectively.

We compare Mol-PECO with the baseline models (Table 2): the conventional GCN of graph representations, including the adjacency matrix and the CM, and the classifiers of fingerprint representations, including Mordreds features (mordreds)\citep{moriwaki2018mordred}, bit-based fingerprints (bfps)\citep{rogers2010extended}, and count-based fingerprints (cfps)\citep{rogers2010extended}. 
Conventional classifiers include $k$-Nearest Neighbor (KNN), random forest (RF), and gradient boosting (GB). In the fingerprint methods, we first handle the problem of imbalanced label distribution with Synthetic Minority Over-sampling Technique (SMOTE)\citep{chawla2002smote}, and then perform the classification. 
Mol-PECO outperforms the baselines in three out of six evaluation metrics (Table 2), with AUROC improved from $0.761$ (cfps-KNN) to $0.813$, AUPRC improved from $0.144$ (mordreds-RF) to $0.181$, and accuracy improved from $0.780$ (Coulomb-GCN) to $0.808$.
Notably, Mol-PECO can balance AUROC ($0.813$) and AUPRC ($0.181$) whereas the ML method
(cfps-KNN) with the highest AUROC of $0.761$ has a low AUPRC of $0.057$ and one (cfps-RF) with the
highest AUPRC of $0.144$ shows low AUROC of $0.723$.
Thus, Mol-PECO boosts the predictability of QSOR.

\begin{table}[htbp]
\label{tab3}
\centering  
\begin{threeparttable}  
  \caption{Performance comparison of Mol-PECO with baseline methods, GCN, and Coulomb-GCN.  The highest scores are shown in bold. The runner-ups are shown with underlines.}
  {\scriptsize \normalsize
  \setlength\tabcolsep{4pt}
  \begin{tabular}{ccccccc}
  \toprule  
  {Baseline\tnote{1}} & {AUROC\tnote{2}} & {AUPRC\tnote{2}} & {Precision\tnote{2}} & {Recall\tnote{2}} & {Specificity\tnote{2}} & {Accuracy\tnote{2}}\\
  \midrule
    cfps-KNN & 0.761 & 0.057 & 0.065 & 0.760 & 0.764 & 0.762\\
    bfps-KNN & 0.758 & 0.055 & 0.064 & 0.748 & 0.769 & 0.759\\
    mordreds-KNN & 0.729 & 0.052 & 0.062 & 0.676 & 0.783 & 0.730\\
    mordreds-RF & 0.723 & 0.144 & \underline{0.241} & 0.483 & \textbf{0.964} & 0.723\\
    cfps-RF & 0.689 & 0.137 & \textbf{0.258} & 0.418 & 0.961 & 0.690\\
    bfps-RF & 0.671 & 0.119 & 0.227 & 0.381 & \underline{0.962} & 0.672\\
    mordreds-GB & 0.725 & 0.126 & 0.220 & 0.499 & 0.951 & 0.725\\
    cfps-GB & 0.701 & 0.120 & 0.210 & 0.453 & 0.950 & 0.702\\
    bfps-GB & 0.687 & 0.111 & 0.193 & 0.428 & 0.948 & 0.688\\
    \hline 
    adjacency-GCN & 0.678 & 0.111 & 0.079 & \textbf{0.827} & 0.625 & 0.726\\
    Coulomb-GCN & 0.759 & 0.143 & 0.089 & 0.816 & 0.744 & 0.780\\
    \hline
    Mol-PECO-sym & \underline{0.796} & \underline{0.153} & 0.088 & 0.817 & 0.787 & \underline{0.802} \\
    Mol-PECO-asym & \textbf{0.813} & \textbf{0.181} & 0.104 & \underline{0.819} & 0.797 & \textbf{0.808}\\
    
    \bottomrule
\end{tabular}   
\begin{tablenotes}  
\item[1] Baseline includes conventional classifiers of fingerprint representations and graph convolutional networks (GCN) of molecular graphs. Fingerprint representations include count-based fingerprints (cfps),  bit-based fingerprints (bfps), and Mordreds features (mordreds). The conventional classifiers include $k$-Nearest Neighbor (KNN), random forest (RF), and gradient boosting (GB). Molecular graph representations include adjacency matrix (adjacency-GCN) and Coulomb matrix (Coulomb-GCN). Mol-PECO-sym refers to the performances with LPE of raw Coulomb matrix. Mol-PECO-asym refers to the performances with LPE of asymmetrically normalized Coulomb matrix.

\item[2] The evaluation metrics are calculated in testing set. 
\end{tablenotes}  
}  
\end{threeparttable}
\end{table}

\subsection{Learned odor space by Mol-PECO}
To investigate the learned structure of multiple odors in relation with descriptors, we perform dimensionality reduction over the output of Mol-PECO's penultimate layer to build the latent odor space and evaluate it at global and local scales. 
At the global scale, we inspects how appropriately the clusters of odors in this latent space represent the information of descriptors, while, at the local scale, we examines whether individual molecules can possess a set of odor descriptors similar to those of nearby molecules or not. 

\begin{figure}[htbp]
\centering
\includegraphics[width=0.95\linewidth]{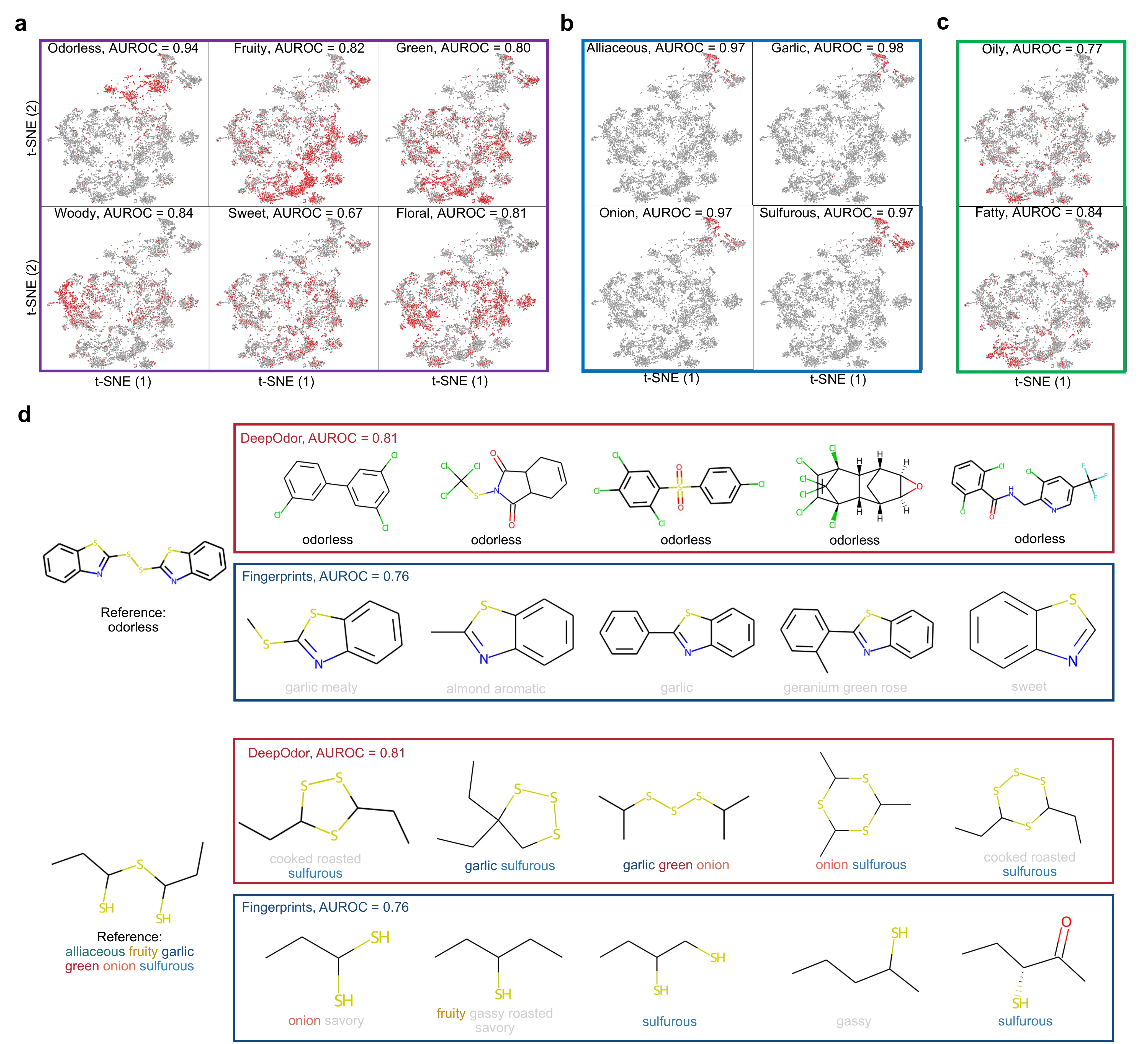}
\caption{The odor space built from Mol-PECO and its global and local properties. 
Global view of the learned odor space with dimensionality reduction by t-SNE is shown in (a), (b), and (c) where each red or gray dot represents one odor molecule with and without the corresponding descriptor, respectively. The value of AUROC in each panel is the AUROC of the corresponding descriptor.
(a) The descriptors with most molecules (`odorless', `fruity', `green', `woody', `sweet', and `floral'), (b) those for the sulfurous compounds (`alliaceous', `garlic', and `onion'), and (c) those with similar semantic meaning (`oily' and `fatty'). 
(d) Local view of learned odor space investigated with nearest neighbor retrieval of reference odorless and odorant molecules. AUROC in (a), (b), and (c) refers to the AUROC of single odor descriptor. AUROC in (d) refers to the unweighted AUROC of 118 odor descriptors.}
\label{fig:fig5}
\end{figure}

For global structure, we analyze the distribution of the high-frequency descriptors, the structure-correlated descriptors, and the synonymy descriptors. For the high-frequency descriptors, we evaluate `odorless', `fruity', `green', `woody', `sweet', and `floral', which are the top six most assigned descriptors. 
The odorless molecules show a high AUROC value (AUROC = 0.94) and cluster far away from other molecules (Figure \ref{fig:fig5}a, top left panel), verifying Mol-PECO's ability to distinguish odorant and odorless molecules. 
Descriptor `woody' localizes on the clusters at the middle of the odor space, whereas descriptors of `fruity', `floral', and `green' are distributed across multiple clusters over the learned odor space (Figure \ref{fig:fig5}a), indicating that 'woody' is a more different characteristic than `fruity', `floral', and `green'.
As shown in Figure \ref{fig:fig2}c, these three descriptors tend to co-occur. 
In the odor space, we can find clusters for different pairs of them, meaning that Mol-PECO can disentangle the differences between `fruity'-`floral', `floral'-`green', and `fruity'-`green'.
The descriptor `sweet' achieves a low AUROC score (AUROC = $0.67$) and the associated molecules spread over the space without a specific pattern, presumably reflecting its polysemous nature and strong association with taste.

For structure-correlated descriptors , molecules associated with `alliaceous', `garlic', or `onion' appear in the cluster of `sulfurous' compounds (Figure \ref{fig:fig5}b) and show a very high score (AUROC = $0.97-0.98$), in accordance with previous studies about sulfur compounds' olfactory descriptions\citep{block1992organosulfur, mikekus2020health}. 
For synonymy descriptors, molecules of ‘oily' and ‘fatty' are clustered into neighbors, validating their similar semantics (Figure \ref{fig:fig5}c). 

For local structure, we investigate one odorless molecule (triphenylphosphane) and one odorant molecule (1-(1-sulfanylpropylsulfanyl)propane-1-thiol with descriptors of `alliaceous', `fruity', `garlic', `green', `onion', and `sulfurous') as the examples (Figure \ref{fig:fig5}d). 
We compare the top-5 nearest molecules searched by Mol-PECO's embedding and bfps. 
The top-5 molecules of Mol-PECO and bfps are calculated with cosine similarity and Tanimoto similarity, respectively. 
For the odorless molecule, Mol-PECO retrieves all neighbors with the odorless descriptor, but bfps retrieves no molecule.
Notably, the molecules fetched by Mol-PECO possess different substructures compared with the reference (e.g., all with C-Cl bond and top-2 / 3 / 5 with carbonyl functional group).
For the odorant molecule, Mol-PECO retrieves all neighbors with shared descriptor, and bfps retrieves four. 
Moreover, all the molecules retrieved by bfps are open-chain structured, the same with the reference. 
In contrast, Mol-PECO retrieves quite different structured molecules, four of which are cyclic molecules.
Both examples would indicate Mol-PECO's promising potential in decoding molecules with different structures but identical smells.

\section{Discussion}
In this work, we develop Mol-PECO for predicting human olfactory perception from molecular structures. 
We handle this QSOR problem by improving the graph-neural-network-based approach from two aspects: the molecular representation and the graph modeling method. 
For molecular representation, we use CM, which is fully-connected and contains 3D conformer information (3D coordinates and charges of atoms), instead of adjacency matrix. 
For graph modeling, we use the positional encoding from the eigenfunctions of Laplacian to make up for GCN's deficiency in directional modeling. 
Mol-PECO improves the olfactory perception prediction in two stages: starting from adjacency-GCN (AUROC of $0.678$) to Coulomb-GCN (AUROC of $0.759$) by introduction of CM, following by from Coulomb-GCN (AUROC of $0.759$) to Mol-PECO with eigenfunctions of the symmetric (AUROC of $0.796$) and asymmetric Laplacian (AUROC of $0.813$).

These results indicate that improvement of the expressive power of GCN can greatly contribute to learning of non-trivial relationships between molecular structures and olfactory descriptors (labels) in the QSOR problem. 
Although extensions of NN architectures are typically accompanied by an increased cost of learning, the limited size of odorant molecules up to $400$ daltons in molecular weight enables the practical application of such extended architectures.
Thus, we believe that the QSOR problem provides a good real-world task to test and demonstrate the effectiveness of advanced architectures. 
In particular, pre-training of graph representation using unlabeled data may lead to further improvements in accuracy of QSOR prediction\citep{DBLP:journals/corr/abs-2012-11175, DBLP:journals/corr/abs-1905-12265}.
In addition, as we found by experiments, employment of asymmetric Laplacians might also contribute to further technical advancements.

However, the QSOR problem suffers from ambiguity in the labels (descriptors) assigned to each molecule and also from low objectivity and consistency of the labeling conducted by few specialists\citep{trimmer2019genetic, keller2007genetic}. 
In particular, individuals can have a quite different sense to highly ambiguous descriptors such as `sweet'\citep{keller2007genetic}.
Thus, consistent labeling by individuals requires a certain amount of training, resulting in the difficulty to increase the amount of data.
Moreover, for the prediction of mixed odors, which is important for applications, data acquisition is prohibitive due to the huge number of possible combinations\citep{meister2015dimensionality}.
In the future, it will be important to apply the approach developed in this paper to the prediction of more objective and systematically measurable outputs, such as the chemical properties of odorants\citep{pannunzi2019odor}, the response of olfactory receptors\citep{bhandawat2005elementary, mainland2015human}, and the response of neural activity\citep{haddad2010global, lapid2011neural}. 
Such extensions will lead to more comprehensive and data-oriented understanding of chemical information coding in the olfactory system.

\section{Materials and methods}
\subsection{Coulomb matrix and its normalization}
As a molecular representation, Coulomb matrix is calculated mainly based on Coulomb replusion force as follows:
\begin{equation}
    C_{i j}=\left\{\begin{array}{ll}
0.5 Z_{i}^{2.4} & \text { for } i=j \\
\frac{Z_{i} Z_{j}}{\left|\mathbf{R}_{i}-\mathbf{R}_{j}\right|} & \text { for } i \neq j
\end{array}\right.,
\end{equation}
where $C_{ij}$ refers to the entry in $i^{th}$ row and $j^{th}$ column, $Z_i$ refers to the atomic charge, and $R_i$ refers to the relative coordinates.

Considering the gap of minimal and maximal entries in Coulomb matrix, we perform 2 preprocessing methods to handle it, including matrix-wised Frobenius normalization and minmax normalization as follows:
\begin{equation}
    \begin{aligned}
    ||C||_F &= \sqrt{\sum_{i = 1}^{n}\sum_{j = 1}^{n}C_{ij}^2},\\
    C_{max} &= \max_{i, j}{C_{ij}},\\
    C_{min} &= \min_{i, j}{C_{ij}},\\    
    \end{aligned}
\end{equation}
where $||C||_F$ refers to the normalization term of Frobenius normalization, $C_{max}$ and $C_{min}$ refer to the max and min term in minmax normalization. With the normalization term, we calculated the preprocessed Coulomb matrix as follows:
\begin{equation}
    \begin{aligned}
    C^F &= C / ( ||C||_F + \epsilon),\\
    C^M &= (C - C_{min}) / (C_{max} - C_{min} + \epsilon), \\
    \end{aligned}
\end{equation}
where $C^F$ and $C^M$ refer to the Frobenuis- and minmax- normalized matrix, and $\epsilon$ equals to ${10}^{-9}$.

\subsection{GCN of Coulomb matrix for multi-label classification}
We build Coulomb-GCN based on Coulomb matrix with three modules: 1) atom embedding updating, 2) molecular embedding extraction, and 3) multi-label classification. In atom embedding updating, we use GCN with the residual mechanism to learn the molecular embedding as follows:
\begin{equation}
    H^{(l)}=\sigma(XH^{(l-1)}W_{graph}^{(l-1)})+H^{(l-1)}W_{linear}^{(l-1)},
\end{equation}
where $H^{(l)} \in R^{n \times d}$ refers to the updated atom embedding in $l^{th}$ layer, $X \in R^{n \times n}$ refers to Coulomb matrix, $H^{(l-1)} \in R^{n \times h}$ refers to the updated atom embedding in ${l-1}^{th}$ layer,
$W_{graph}^{(l-1)} \in R^{h \times d}$ refers to the parameters in GCN, $W_{linear}^{(l-1)} \in R^{h \times d}$ refers to the parameters in the linear transformation of $H^{(l-1)}$ for residual mechanism, $H^{(0)}$ refers to the random initialized atom embedding, $n$ refers to the number of atom in molecule, $d$ refers to the dimension of embedding in $l^{th}$ layer, $h$ refer to the dimension of embedding in ${l-1}^{th}$ layer, and $\sigma$ refers to SELU activation function\citep{DBLP:journals/corr/KlambauerUMH17}. 
In molecular embedding extraction, we use the sum-pooling function as follows:
\begin{equation}
    m_i = \sum_{j\in [n]} H_{ji}^{(l)},
\end{equation}
where $m_i$ refers to the $i^{th}$ entry in the molecular embedding of molecule $m$. In multi-label classification, we use fully-connected layers as follows:
\begin{equation}
    y = \sigma(mW_{clf}),
\end{equation}
where $m \in R^{1 \times d}$ refers to the learned molecular embedding, $W_{clf} \in R^{d \times o}$ refers to the parameters in the fully-connected layer, and $o$ refers to the number of odor descriptors. Coulomb matrix and Coulomb-GCN have been implemented by Python (version 3.7.4) with deepchem (version 2.6.1) and pytorch (version 1.12.1).

\subsection{Directional graph modeling by Laplacian for multi-label classification}
Mol-PECO differs from Coulomb-GCN only in the atom embedding updating module, where we build the directional graph modeling with learned positional encoding (LPE).

The Laplacian matrix ($L^1$) of Coulomb matrix ($X$) is calculated as follows:
\begin{equation}
    L^1 = D - X,
\end{equation}
where $D$ refers to the degree matrix. In LPE with eigen-decomposition of Laplacian, symmetrical Laplacian matrix is calculated for further spectral decomposition as follows:
\begin{equation}
    \begin{aligned}
    L^2 &= I - D^{-1/2}XD^{-1/2}\\
        &= D^{-1/2}DD^{-1/2} - D^{-1/2}XD^{-1/2}\\
        &= D^{-1/2}(D-X)D^{-1/2}\\
        &= D^{-1/2}L^1D^{-1/2}.\\
    \end{aligned}
\end{equation}
where $I$ refers to the identity matrix. With the equations, $L_2$ is calculated with $L_1$ divided by $\sqrt{d_{ii}d_{jj}}$. Let $d_{ii} = \sum_{j=0}^{i-1}{a_{ij}} + \sum_{j=i+1}^{n}{a_{ij}} + a_{ii}$, then the Laplacian is calculated as follows:
\begin{equation}
    \begin{aligned}
    L^1_{ij} &= \sum_{j=0}^{i-1}{a_{ij}} + \sum_{j=i+1}^{n}{a_{ij}} + a_{ii} - a_{ii}, i = j,\\
    L^1_{ij} &= -a_{ij}, i \neq j,\\   
    \end{aligned}
\end{equation}
which indicates setting the diagnoal entries zeros or not has no influence on $L^1$ and $L^2$ as $L^2 = D^{-1/2}L^1D^{-1/2}$.

Minimization of quadratic form on graphs acts as the cost function of link/edge prediction and captures the global structural information. Naturally, the eigen-decompostion of Laplacian provides the solutions as follows:
\begin{equation}
    f^* = \min\limits_{f}fL^2f^T = 1/2\sum_{i, j}X(i, j){(f_i-f_j)}^2,
\end{equation}
where $f^*$ refers to the eigenvectors of $L$.
With Laplacian, Mol-PECO updates the positional encoding in an atom-wise manner proposed in Spectral Attention Network\citep{kreuzer2021rethinking}. Mol-PECO learns LPE in an atom-by-atom manner. Specifically, Mol-PECO first performs the linear transformation of eigenvalues and eigenvectors in a single atom, learns the positional information by Transformer, and then extracts the atom positional encoding by sum-pooling as follows:
\begin{equation}
    \begin{aligned}
        W &= \Lambda W_0,\\
        K &= Transformer(W),\\
        e_i &= \sum_{j \in n}K_{ji}, 
    \end{aligned}
\end{equation}
where $\Lambda \in R^{p \times 2}$ refers to the $p$ eigenvalues and eigenvectors for a single atom, $W_0 \in R^{2 \times d}$ refers to the parameters of the linear transformation in LPE, $K \in R^{p \times d}$ refers to the updated embedding with Transformer of $W \in R^{p \times d}$, and $e_i$ refers to the $i^{th}$ entry of single atom embedding (vector-shaped representation). Graph Laplacian and Mol-PECO have been implemented by Python (version 3.7.4) with deepchem (version 2.6.1) and PyTorch (version 1.12.1).
\subsection{Loss functions}
For both Coulomb-GCN and Mol-PECO, we build the classification loss with the binary cross-entropy loss function and a logarithm regularization term as follows:
\begin{equation}
    \begin{aligned}
      l^i(y_{true}^i, y_{pred}^i) &= BCE(y_{true}^i, y_{pred}^i ) + \lvert log⁡(y_{pred}^i + \epsilon) - log⁡(y_{true}^i + \epsilon) \rvert,\\
      l(y_{true}, y_{pred}) &= \frac{1}{o} \sum_{i \in [o]} w_il^i(y_{true}^i, y_{pred}^i),
    \end{aligned}
\end{equation}
where $y_{true}^{i}$ and $y_{pred}^{i}$ refer to the ground truths and predictions of $i^{th}$ odor descriptor, $l^i$ refers to the loss function of $i^{th}$ odor descriptor, BCE refers to the binary cross-entropy function, $w_i$ refers to $1-n_{pos}^i/n_{tot}$, $n_{pos}^i$ refers to the number of positive samples in $i^{th}$ descriptor, $n_{tot}$ refers to the number of total samples. The training process has been implemented by Python (version 3.7.4) with PyTorch (version 1.12.1).

\subsection{Molecular descriptors}
We include three classical molecular descriptors in this work as the baseline molecular representations, including Mordred features (mordred)\citep{moriwaki2018mordred}, bit-based Morgan fingerprints (bfps)\citep{rogers2010extended}, and count-based Morgan fingerprints (cfps)\citep{rogers2010extended}. 
Mordred calculates about 1825 features, including 214 2D and 1611 3D features. 
Both bfps and cfps encode the molecule's topological environments (molecular fragments), which indicate the presence of atoms and functional groups, into a vector. 
Specifically, bfps encodes presence or absence of the molecular fragments as a binary information, while cfps encodes the number of atom/functional-group in the topological environment. 
The calculation of molecular descriptors has been implemented by Python (version 3.7.4) with mordred (version 1.2.0) and rdkit-pypi (version 2022.3.4).

\subsection{Machine learning of molecular descriptors}
Machine learning of molecular descriptors used in this work as baselines includes: 1) Synthetic Minority Over-sampling Technique (SMOTE)\citep{chawla2002smote} for handling the imbalanced label distribution, and 2) multiple binary classifiers for predicting multiple odor descriptors. For SMOTE, it performs minority sampling by generating new minority instances to expand the number of the minority class. For binary classifiers, we use K- Nearest Neighbor classifier (KNN), random forest classifier (RF), and gradient boosting classifier (GB). The KNN acts as a non-parametric classifier and predicts the label by voting from neighbors. RF acts as the ensemble learning of decision trees with sample bagging to decrease the variance of model and feature bagging to decrease the correlation among decision trees. GB builds multiple weak learners to minimize the differences between the true label and the predicted value by performing gradient decent. The procedures have been implemented by Python (version 3.7.4) with imblearn (version 0.9.0) and scikit-learn (version 1.0.2).

\subsection{Acknowledgments}
We thank the suggestions and discussion with all colleagues in laboratory for quantitative biology, IIS, Tokyo University and Takahiro G. Yamada, Yusuke Shibuya, and Mamoru Tomiyama in Funahashi-lab, Keio University. 
We also thank Yusuke Ihara and Chiori Ijichi for helpful comments.
The work is supported by China Scholarship Council (CSC) Grant 202106230237 to M. Z., JSPS KAKENHI Grant Numbers 19H05799 to T.J.K., and JST CREST Grant Number JPMJCR2011 to A.F. and T.J.K. 
\bibliography{main}
\bibliographystyle{unsrtnat}

\appendix
\section{Appendix}

\subsection{Parameter tuning of Mol-PECO}
We optimize Mol-PECO with different number of 
Transformer layers in learned positional encoding and choose $4$ as the optimized parameter (Table 3).

\begin{table}[htbp]
\label{tab4}
\centering  
\begin{threeparttable}  
  \caption{Scores of Mol-PECO with different number of Transformer layers and the layer number optimization. The highest scores are shown in bold.}  
  {\scriptsize \normalsize
  \setlength\tabcolsep{4pt}
  \begin{tabular}{ccccccc}
  \toprule  
   {\#Transformer layers} & {AUROC\tnote{1}} &	{AUPRC\tnote{1}} & {Precision\tnote{1}} & {Recall\tnote{1}} & {Specificity\tnote{1}} & {Accuracy\tnote{1}}\\
  \midrule
    1 & 0.809 & 0.170 & 0.103 & 0.826 & 0.789 & 0.807\\
    2 & 0.798 & 0.147 & 0.100 & 0.817 & 0.779 & 0.798\\
    3 & 0.807 & 0.163 & 0.090 & \textbf{0.831} & 0.779 & 0.805\\
    4 & \textbf{0.813} & 
    \textbf{0.181} & \textbf{0.104} & 0.819 & 
    \textbf{0.797} & \textbf{0.808}\\
    5 & 0.807 & \textbf{0.181} & 0.089 & 0.819 & 0.780 & 0.800\\
    6 & 0.802 & 0.162 & 0.093 & 0.814 & 0.790 & 0.802\\
    \bottomrule
\end{tabular}   
\begin{tablenotes}
\item[1] The evaluation metrics are calculated with the validation set. 
\end{tablenotes}  
}  
\end{threeparttable}  
\end{table}

\begin{figure}[htbp]
\centering
\includegraphics[width=0.95\linewidth]{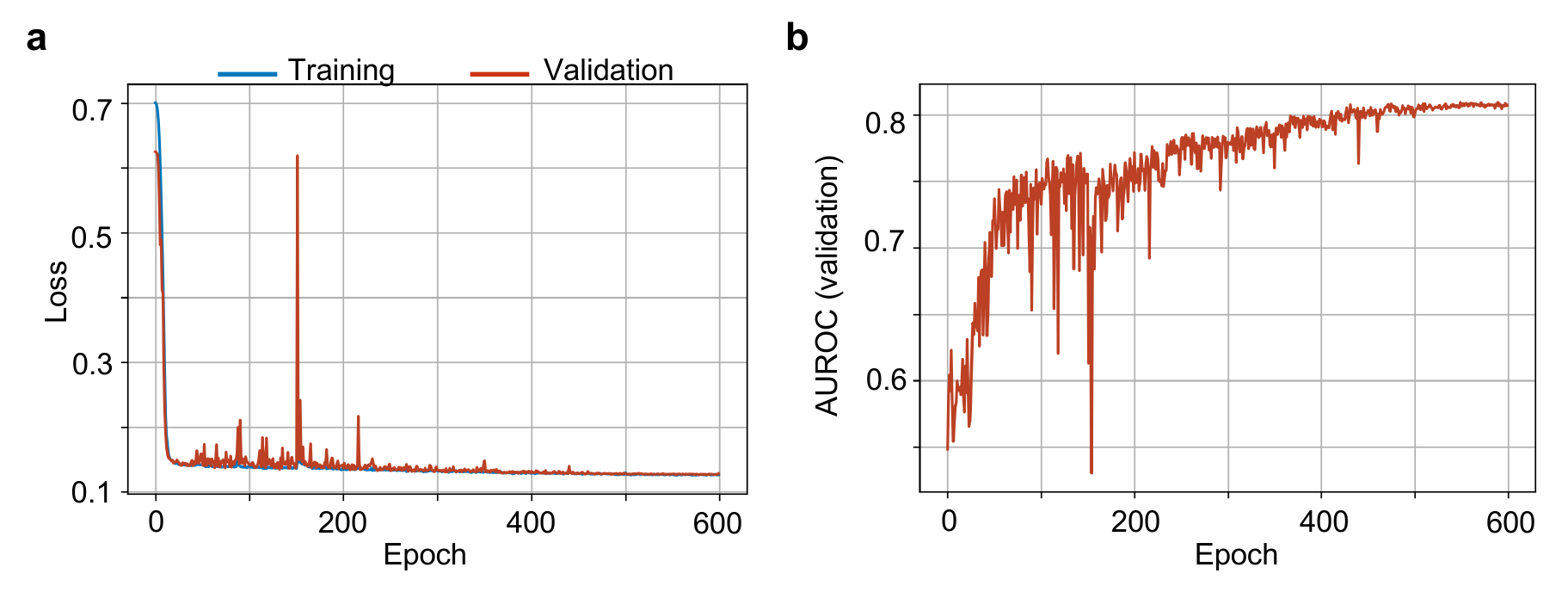}
\caption{The training process with the optimized parameter. (a) The unweighted loss during training. (b) The unweighted AUROC during training.}
\label{fig:figs1}
\end{figure}

The training process of the optimized parameter (the number of Transformer layer = $4$) is monitored by the loss curves (Figure \ref{fig:figs1}a). The decreased trends of the loss curves are consistent, indicating that Mol-PECO has found the right bias/variance tradeoff in the training and validation sets. Mol-PECO chooses the checkpoint of $554$ epochs with minimal loss of $0.140$ as the final model for testing (Figure \ref{fig:figs1}b). 

\begin{figure}[htbp]
\centering
\includegraphics[width=0.95\linewidth]{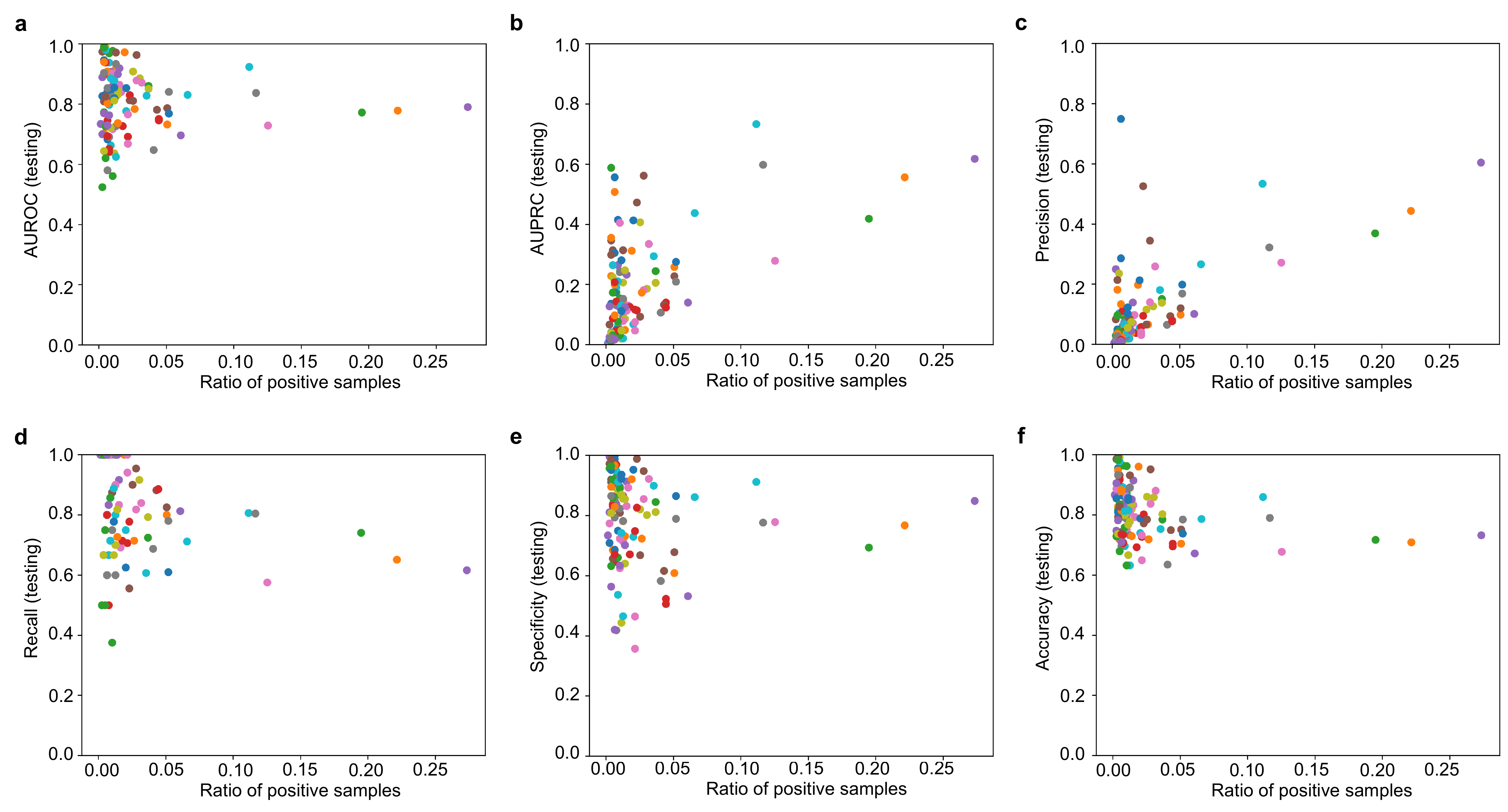}
\caption{The detailed performances of Mol-PECO in 6 evaluation metrics, including (a) AUROC, (b) AUPRC, (c) Precision, (d) Recall, (e) Specificity, and (f) accuracy. The x-axis refers to the ratio of positive samples. Each dot refers to one odor descriptor.}
\label{fig:figs2}
\end{figure}

After training and validation, we obtain the detailed performances of $118$ odor descriptors in $6$ evaluation metrics (Figure \ref{fig:figs2}).

\subsection{Supplementary materials for the learned odor space}
\begin{figure}[htbp]
\centering
\includegraphics[width=0.95\linewidth]{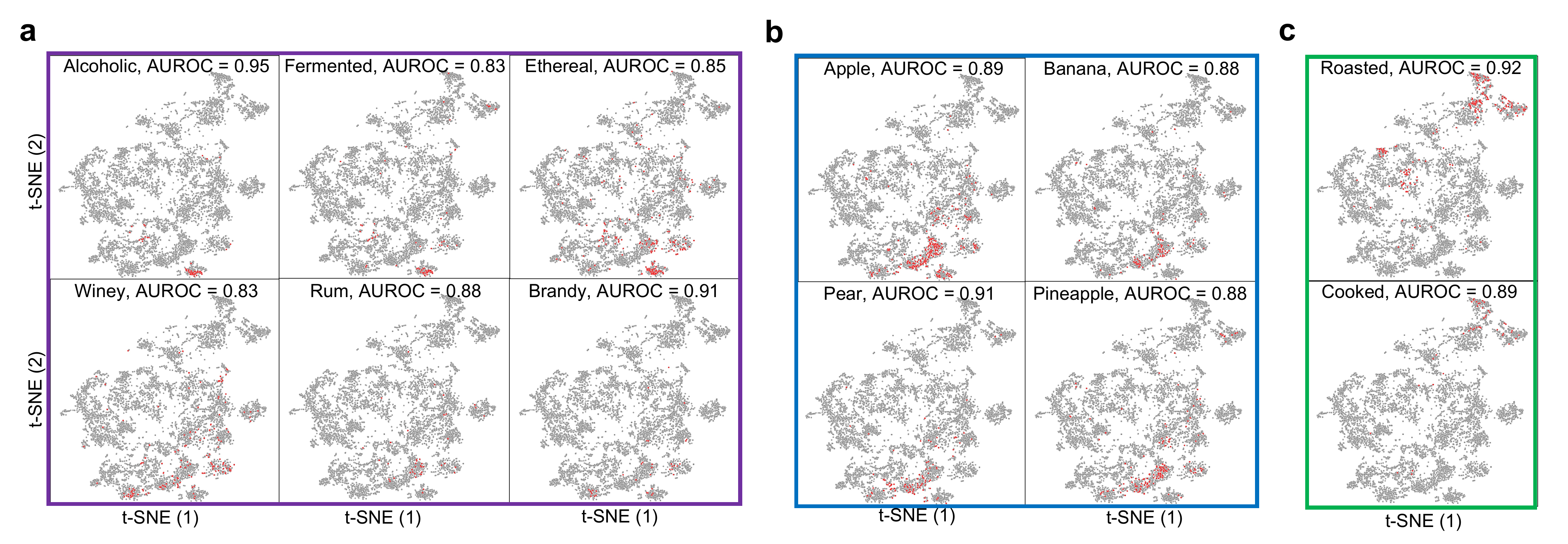}
\caption{Global view of learned odor space with dimensionality reduction by t-SNE in (a) the alcohol-related descriptors (`alcoholic', `fermented', `ethereal', `winey', `rum', and `brandy'), (b) the fruit-related descritors (`apple', `banana', `pear', and `pineapple'), and (c) the molecules with similar semantic meaning (`roasted' and `cooked').}
\label{fig:figs3}
\end{figure}

Despite of the high-frequency descriptors, the structure-correlated descriptors, and the synonymy descriptors, we also investigate the alcohol-related descriptors, the fruit-related descriptors, and the synonym descriptors in Figure \ref{fig:figs3}. For alcohol-related descriptors, `ethereal' and `winey' possess the most molecules, leading to an obvious cluster in nearby locations (Figure \ref{fig:figs3}a). For the fruit-related descriptors, `apple', `banana', `pear', and `pineapple' are clustered together (Figure \ref{fig:figs3}b). For synonymy descriptors, partial ‘roasted' and full ‘cooked' are clustered into neighbors, validating their similar semantics (Figure \ref{fig:figs3}c). 

\end{document}